\documentclass[lettersize,journal]{IEEEtran}
\usepackage{amsmath,amsfonts}
\usepackage{algorithm}
\usepackage{array}
\usepackage[caption=false,font=normalsize,labelfont=sf,textfont=sf]{subfig}
\usepackage{textcomp}
\usepackage{stfloats}
\usepackage{url}
\usepackage{verbatim}
\usepackage{graphicx}
\usepackage{cite}

\usepackage{algorithm,algpseudocode}

\usepackage{amsmath}
\usepackage{booktabs}
\usepackage{array,caption}
\usepackage{tabularx}
\usepackage{lipsum} 
\usepackage{multirow}
\usepackage{graphicx}
\usepackage{color}
\usepackage{amsfonts}
\usepackage{relsize}
\usepackage{caption}
\usepackage[normalem]{ulem}
\useunder{\uline}{\ul}{}
\begin{document}
\title{Mamba4KT:An Efficient and Effective Mamba-based Knowledge Tracing Model}

\author{Yang Cao, and Wei Zhang
\IEEEcompsocitemizethanks{
\IEEEcompsocthanksitem Yang Cao, Jun Wang and Wei Zhang are with the School of Computer Science and Technology, East China Normal University, Shanghai, China. E-mail: \{caoyang99775, zhangwei.thu2011\}@gmail.com.

}
}

\maketitle

\begin{abstract}
Knowledge tracing (KT) enhances student learning by leveraging past performance to predict future performance. Current research utilizes models based on attention mechanisms and recurrent neural network structures to capture long-term dependencies and correlations between exercises, aiming to improve model accuracy. Due to the growing amount of data in smart education scenarios, this poses a challenge in terms of time and space consumption for knowledge tracing models. However, existing research often overlooks the efficiency of model training and inference and the constraints of training resources. Recognizing the significance of prioritizing model efficiency and resource usage in knowledge tracing, we introduce Mamba4KT. This novel model is the first to explore enhanced efficiency and resource utilization in knowledge tracing. We also examine the interpretability of the Mamba structure both sequence-level and exercise-level to enhance model interpretability. Experimental findings across three public datasets demonstrate that Mamba4KT achieves comparable prediction accuracy to state-of-the-art models while significantly improving training and inference efficiency and resource utilization. As educational data continues to grow, our work suggests a promising research direction for knowledge tracing that improves model prediction accuracy, model efficiency, resource utilization, and interpretability simultaneously.
\end{abstract}


\section{Introduction}

\graphicspath{{pdfs/}}
\begin{figure}[h!]
    \centering
    \includegraphics[width=8cm]{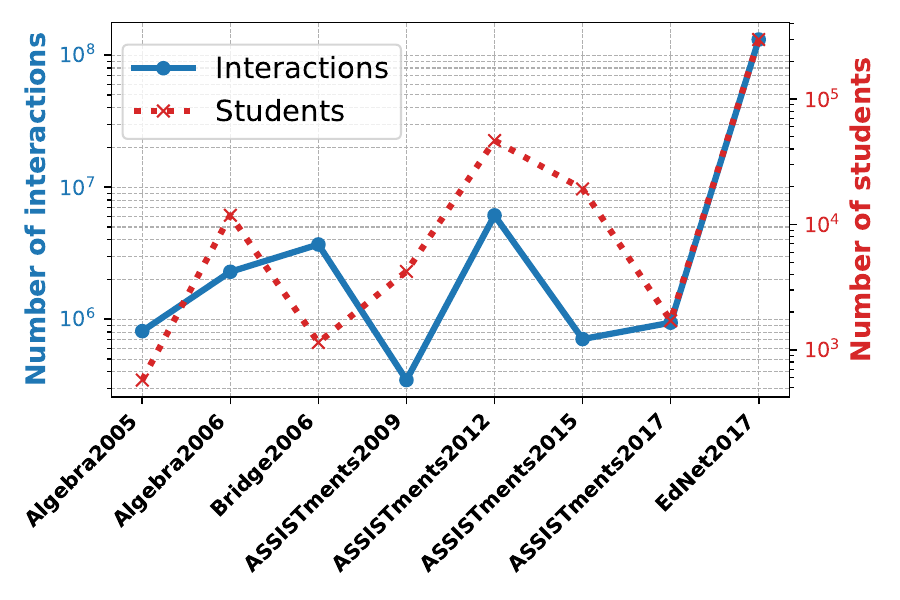}
    \caption{Trends in data volume for education scenarios.}
    \label{fig1}
\end{figure}

\graphicspath{{pdfs/}}
\begin{figure}[h!]
    \centering
    \includegraphics[width=8cm]{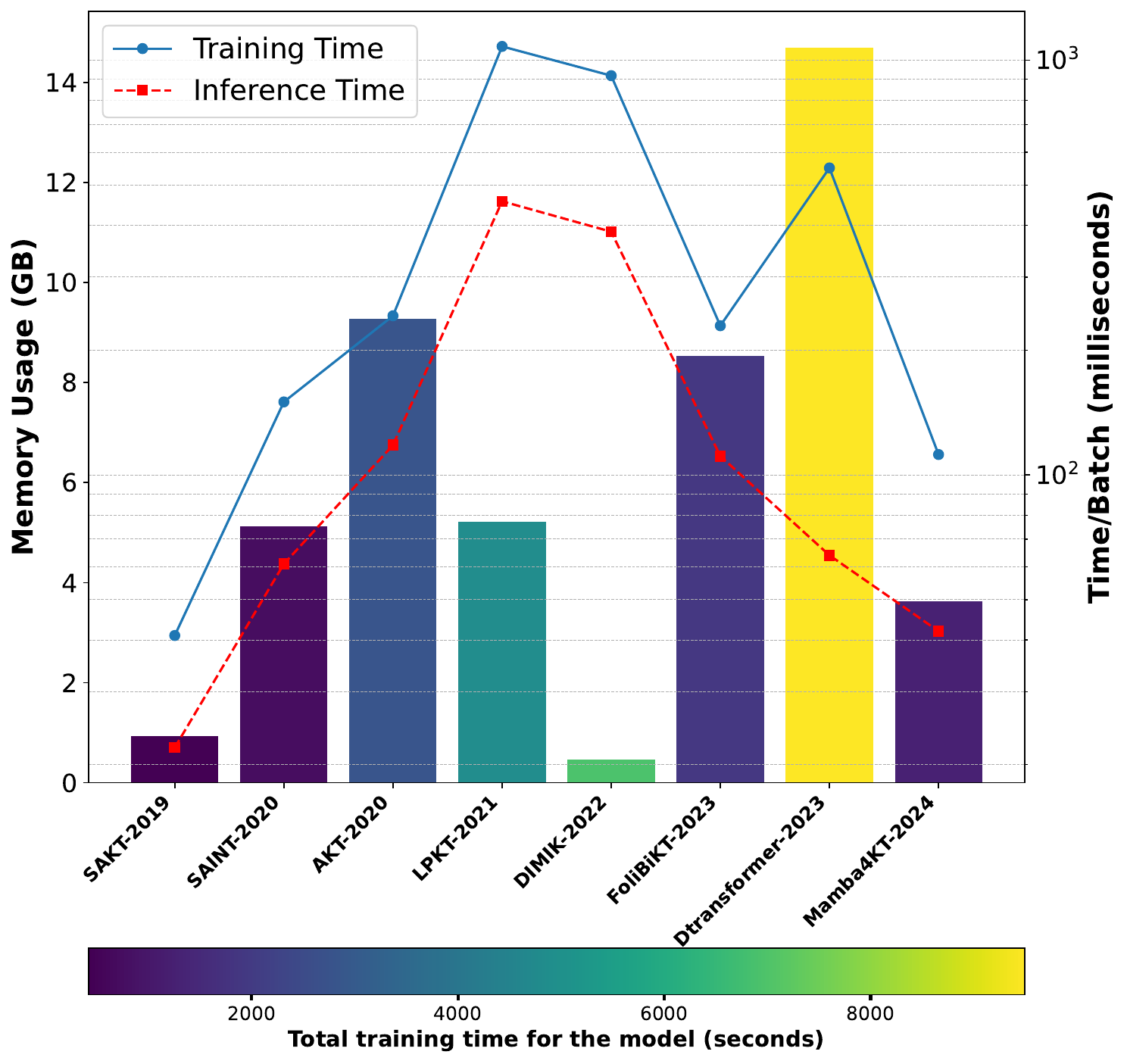}
    \caption{Model performance vs. resource consumption, where x-axis lists various KT models, left y-axis shows training resource usage, right y-axis displays batch training and inference time, bar color indicates total training time.}
    \label{fig2}
\end{figure}
With the popularity of smart education scenarios, there is an explosion of educational data~\cite{gong2020attentional,abdelrahman2023knowledge}. Knowledge Tracing (KT)~\cite{corbett1994knowledge} task, which aims to predict students' knowledge mastery status by analyzing their interaction data, is becoming increasingly important. As shown in Fig.~\ref{fig1}, the 2017 Ednet dataset is about 100 times larger than the 2005 Algebra dataset at the level of the number of interactions and the number of students~\cite{abdelrahman2023knowledge}, and these datasets are only partial datasets of actual scenarios. The increased scale of data volume puts a tougher test on the model's consumption of time and space. 

However, traditional knowledge tracing models mainly focus on improving prediction accuracy and model interpretability~\cite{lpkt,akt,folibikt,dtransformer}. As shown in Fig.~\ref{fig2}, compared to the 2019 SAKT~\cite{sakt} model, models based on attention mechanisms~\cite{attention}, such as SAINT~\cite{saint}, AKT~\cite{akt}, FoliBiKT~\cite{folibikt}, and Dtransformer~\cite{dtransformer}, while improving in prediction accuracy, have higher model complexity due to the fact that these models rely on the attention weights between exercises, which inevitably leads to longer training and inference times, as well as higher high consumption of space resources. On the other hand, models based on RNN~\cite{lstm} structures, such as LPKT~\cite{lpkt} and DIMIK~\cite{dimkt}, are more efficient in terms of space utilization, but the nature of their sequence processing makes parallelization difficult, which affects the efficiency of model training and inference. Therefore, it is a great challenge for knowledge tracing models to pursue prediction accuracy while maintaining efficient training and inference as well as efficient utilization of resources. And how to provide interpretability on that basis.

To address the above challenges, we propose a new knowledge tracing model, Mamba4KT, which not only ensures prediction accuracy and interpretability, but also strikes a balance between time and space consumption. We introduce Mamba~\cite{mamba}, a state-space model~\cite{ssms} that supports parallelized training and linear-time inference, which ensures more efficient training and lower space consumption. In addition, we combine a series of state-of-the-art techniques from traditional knowledge tracing models with Mamba, such as rasch model-based embedding~\cite{irt1,irt2} and feed-forward networks~\cite{attention} which further improves the prediction accuracy in knowledge tracing tasks. We provide a new interpretable approach to the knowledge tracing scenario by introducing the hidden interpretability in the structure of Mamba.
In summary, our contributions lie in four aspects:
\begin{itemize}
\item We introduce a knowledge tracing model Mamba4KT based on selective state space model, which improves the training and inference efficiency and space utilization while ensuring model prediction accuracy.

\item We further combine the advanced techniques in traditional knowledge tracing models with the Mamba structure to improve model prediction while ensuring time and space efficiency.

\item We introduce interpretable methods for selecting state-space models to the knowledge tracing task. Furthermore, our work points to a potential research direction in the field of knowledge tracing that simultaneously improves prediction effectiveness, time and space efficiency, and interpretability performance.

\item Experimental results across various datasets demonstrate that Mamba4KT attains prediction accuracy on par with the current state-of-the-art model while significantly reducing time and space requirements.
\end{itemize}

\section{Related Work}
\subsection{Knowledge tracing}
Early knowledge tracing models are mainly BKT-based models~\cite{bkt} and IRT-based models~\cite{irt1,irt2}. BKT-based models model students' knowledge status through Bayesian inference, which is more interpretive and widely used. IRT-based models use item response theory to analyze students' performance, which is able to quantify students' ability level and the difficulty of the questions, thus providing more personalized learning suggestions. difficulty, thus providing more personalized learning suggestions.

In recent years, knowledge tracing models based on attention mechanisms have gained widespread attention and applications in academia and industry. By introducing the attention mechanism, such models are able to better capture the dependencies of students answering questions at different points in time, which significantly improves the predictive performance of the models.
The SAKT~\cite{sakt} model utilizes a self-attention mechanism to model students' response sequences, effectively capturing the dependencies between students' responses at different points in time.
SAINT~\cite{saint} processes the practice and response embedding sequences separately, and this separation of inputs allows the model to stack the attentional layers multiple times.
AKT~\cite{akt} uses a novel monotonic attentional mechanism to correlate learners' future responses with past responses to assessment questions.
FoliBiKT~\cite{folibikt} allows existing attention-based KT models to easily integrate forgetting behavior by effectively decomposing problem relevance and forgetting behavior.
Dtransformer~\cite{dtransformer} builds from the problem level to the knowledge level to explicitly diagnose learners' knowledge acquisition.

RNN-based knowledge tracing models have also achieved remarkable results.
DKT~\cite{dkt} is a knowledge tracing method based on recurrent neural networks, and the model is able to capture the complex representation of students' knowledge.
The DKT-forget~\cite{dktforget} model adds a forgetting mechanism to DKT, which enables the model to better simulate the process of students' knowledge forgetting.
LPKT~\cite{lpkt} over defines the basic learning units, measures the learning gain and its diversity in depth, and designs learning gates and forgetting gates to differentiate the students' ability to absorb knowledge and simulate the decline of knowledge over time.
                             
Knowledge tracing models based on attentional mechanisms and RNNs show strong advantages in improving prediction accuracy and modeling complex dependencies. However, this part of the knowledge tracing model does not consider the model time complexity and space complexity, and the problem of time and space efficiency of the model becomes more serious as the scale of educational data continues to grow.

\subsection{Selective State Space Models}
The transformer~\cite{attention} architecture has shown strong performance in various tasks. However, this architecture has a huge demand on computational resources, and in order to further reduce the complexity of the attention model, state-space models (SSMs)~\cite{ssms} have received increasing attention in recent years as a possible alternative.

A series of research works are widely used in scenarios such as vision field, medicine field, and time series.
Vision mamba~\cite{vmamba} uses bidirectional Mamba blocks and position embeddings to mark image sequences and compress visual representations, achieving higher performance and improved computation \& memory efficiency compared to established vision transformers like DeiT~\cite{touvron2021training}.
U-Mamba~\cite{umamba} is a novel network for biomedical image segmentation that combines CNNs~\cite{lecun1998gradient} with State Space Sequence Models (SSMs) to address the challenge of capturing long-range dependencies.
SiMBA~\cite{tmamba} is a novel architecture that combines Einstein FFT (EinFFT) for channel modeling with the Mamba block for sequence modeling, addressing stability issues and enhancing performance in State Space Models (SSMs).

\section{Preliminary}
\graphicspath{{pdfs/}}
\begin{figure*}[h!]
    \includegraphics[width=16cm]{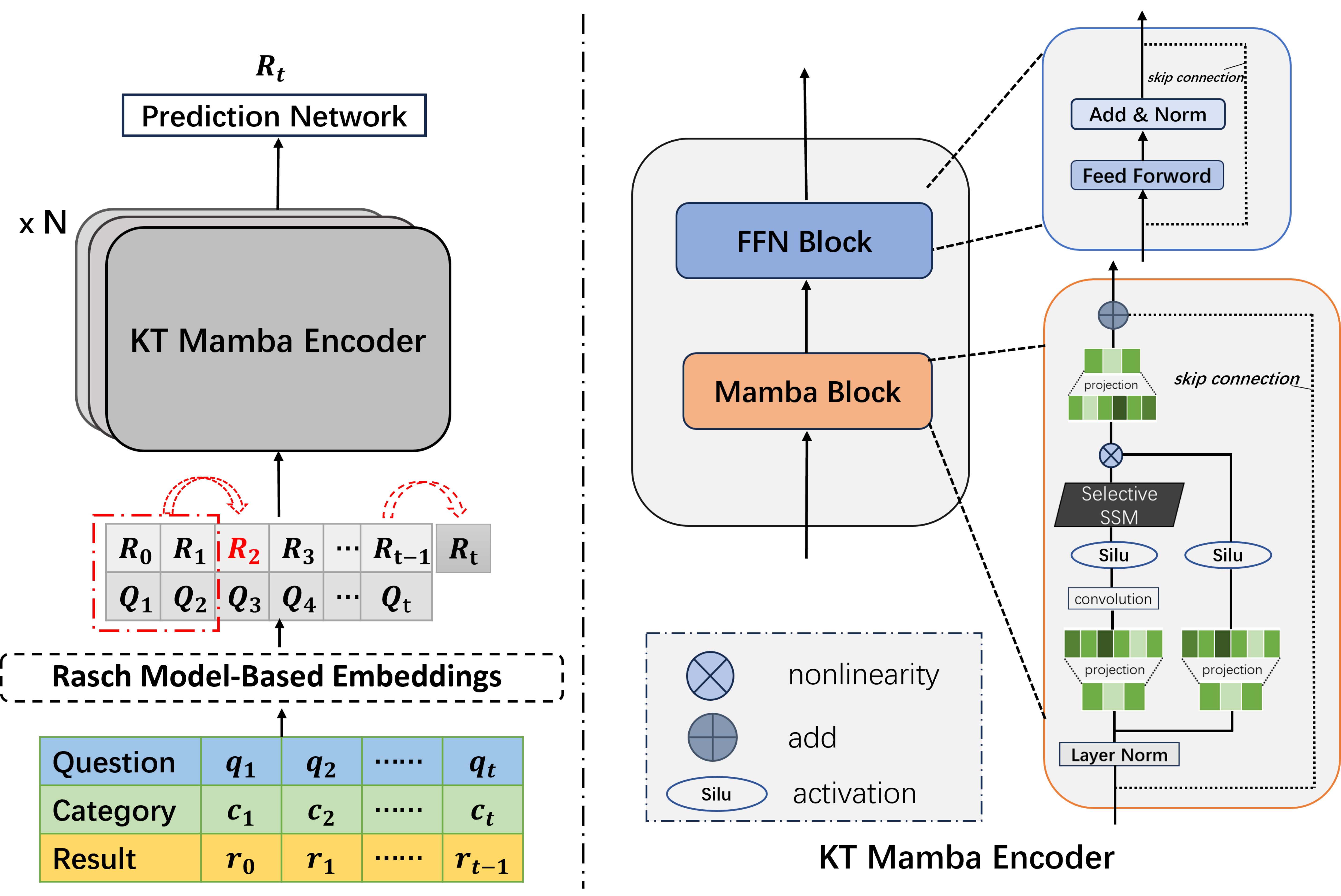}
    \caption{The overview of the proposed framework.}
    \label{framework}
\end{figure*}
\textbf{Problem Formulation} \quad 
In the knowledge tracing task, suppose that there exists a set of questions $Q$, a set of knowledge point categories $C$, and a sequence of $t$ moments of student work $S=\{(q_0,c_0,r_0),(q_1,c_2,r_2),...,(q_{t-1},c_{t-1},r_{t-1})\}$ containing questions, knowledge point categories, and response results, where $q_i\in Q, c_i \in C, r_i \in \{0,1\},0 \leq i \leq t-1$. $(q_i,c_i,0)$ means that the student answered incorrectly when answering question $q_i$ with knowledge point category $c_i$ at moment $i$. On the contrary, $(q_i,c_i,1)$  means that the student answered correctly when answering question $q_i$ with knowledge point category $c_i$ at moment $i$. Mamba4KT predicts students' responses at moment $t$ based on the question sequence $S$ as follows:
\begin{equation}
    \hat{r_{t}} = Mamba4KT(S,q_t,c_t|\theta)
    \label{eq:rt}
\end{equation}
where $\theta$ denotes the parameters for model training, $S$ denotes the information about the sequence of questions done from $0$ to moment $t-1$, $c_t$ denotes the category of knowledge point of the question at moment $t$, $q_t$ denotes the problem at moment $t$, and $\hat{r_{t}}$ denotes the predicted probability of answering correctly at moment $t$.

\textbf{State Space Models} \quad
The State Space Model (SSM) constitutes a framework for sequence modeling that is grounded in linear ordinary differential equations.
It transforms an input sequence $x(t)$ into an output sequence $y(t)$ via an intermediary latent state $h(t)$:
\begin{equation}
    \label{eq:ssm}
    \begin{aligned}
    h^{\prime}(t) & =\boldsymbol{A} h(t)+\boldsymbol{B} x(t), \\
    y(t) & =\boldsymbol{C} h(t),
    \end{aligned}
\end{equation}
where $\boldsymbol{A}$ is a non-learning parameter and $\boldsymbol{B}$ and $\boldsymbol{C} $ are learnable parameters. 

The model is discretized using a zero-order holding basis on top of SSM. It works as follows, each time a discrete signal is received, its value is held until a new discrete signal is received. This process will produce a continuous signal that SSM can use:
\begin{equation}
\label{eq:s6}
\begin{aligned}
h_t & =\overline{\boldsymbol{A}} h_{t-1}+\overline{\boldsymbol{B}} x_t, \\
y_t & =\boldsymbol{C} h_t,\\
\overline{\boldsymbol{A}} & =\exp (\Delta \boldsymbol{A}),\\
\overline{\boldsymbol{B}} & =(\Delta \boldsymbol{A})^{-1}(\exp (\Delta \boldsymbol{A})-\boldsymbol{I}) \cdot \Delta \boldsymbol{B}
\end{aligned}
\end{equation}
where $\Delta$ indicates the length of time the value is held.

Mamba offers several advantages over traditional State-Space Models (SSM). Firstly, it utilizes recurrent SSMs created through discretization. Secondly, it employs HiPPO initialization on matrix $A$ to capture long-range dependencies. Thirdly, Mamba implements a selective scan algorithm to selectively compress information, and finally, it includes a hardware-aware algorithm designed to speed up computation.
\section{Proposed Frameworks}

As shown in Figure~\ref{framework}, we propose a knowledge tracing model based on the selective state-space model, named Mamba4KT. The model consists of three main components: Rasch model embed-based embeddings, Mamba Block, and FFN Block. The Rasch model embed-based embeddings component is used to construct embeddings for questions and knowledge. The Mamba Block and FFN Block replace the traditional attention mechanism in knowledge tracing tasks. This replacement not only captures dependencies between response sequences but also ensures high efficiency in model training and inference, as well as low memory consumption. Additionally, the Mamba Block captures interpretable information at both the sequence level and the question level.
In the rest of this section, we elaborate on the details of the framework.

\subsection{Rasch model embed-based Embeddings}
To provide richer information about questions and knowledge points for the Mamba module, we refer to AKT~\cite{akt} and integrate the Rasch model, incorporating the intrinsic difficulty of questions and students' abilities into the information representation.
\begin{equation}
\label{eq:QR}
\begin{aligned}
    {Q_{t}} & = \mathbf{c}_{c_{t}} + \mu_{q_t}\cdot \mathbf{d}_{c_t}, \\
    {R_{t}} & = \mathbf{e}_{r_t} + \mu_{q_t}\cdot \mathbf{f}_{(c_t,r_t)}
\end{aligned}
\end{equation}
where 
$\mathbf{c}_{c_{t}}$  represents the knowledge category embedding for the question, 
and $\mathbf{e}_{r_{t}}$ represents the result embedding,
and $\mu_{q_t}$ is a scalar difficulty parameter controlling the question's deviation from its covered concept, 
and $\mathbf{d}_{c_t}$  is a vector summarizing variations in questions covering the concept, 
and $\mathbf{f}_{(c_t,r_t)}$ denotes variation vectors for knowledge category-result pairings.

\subsection{Mamba Block}

The Mamba block is constructed using the selective state-space layer (S6), Conv1D, and various elementwise operators. Given an input $\hat{x}' := (x_0, \cdots ,x_{t-1})$ it is defined as follows:

\begin{equation}
\label{eq:mamba_block}
\begin{aligned}
\hat{x} &= \text{SiLU( Conv1D( Linear(}\hat{x}'\text{) ) )}, \\
\hat{z} &= \text{SiLU( Linear(}\hat{x}'\text{) )},  \\
\hat{y}'&= \text{Linear}(\text{S6}(\hat{x}) \otimes \hat{z})), \\
\hat{y} &= \text{LayerNorm}(\hat{y}' + \hat{x}')
\end{aligned}
\end{equation}

Relevant study~\cite{hidden_attention} has shown that the Mamba structure contains latent interpretable information, where 
${\alpha}\in \mathbb{R}^{M\times t \times t}$(M indicates the number of output channels of Conv1D in Eq.~\ref{eq:mamba_block}. ) represents the influence weight of the $jth$ exercise on the  $ith$ exercise.
\begin{equation}\label{eq:attn}
    \alpha_{m,i,j} = C_i \Big{(}\Pi_{k=j+1}^i \bar{A}_k \Big{)} \bar{B}_j, 0\leq m < M, 0\leq j < i\leq t-1
\end{equation}

\noindent\textbf{Sequence-level interpretability.}
Interaction weights between sequences of questions of length t are computed by Eq.~\ref{eq:attn} and normalized by Eq.~\ref{eq:exp_global}. $\gamma_{m, i, j}$ denotes the attentional weight between $x_i$ and $x_j$ in the interpretable information provided by Channel $m$.
\begin{equation}
\label{eq:exp_global}
\begin{aligned}
\gamma_{m, i, j} = \frac{\alpha_{m, i, j}}{\sum_{k=0}^{i-1} \alpha_{m, i, k}}
\end{aligned}    
\end{equation}

\noindent\textbf{Exercise-level interpretability.}
The influence weight of the first t-1 exercises on the exercise $x_t$ is calculated by Eq.~\ref{eq:attn} and normalized by Eq.~\ref{eq:exp_exe}. $\gamma_j$ denotes the influence weight of $x_j$ on $x_i$.
\begin{equation}
\label{eq:exp_exe}
\begin{aligned}
    \beta_j    &= \sum_{m=0}^{M-1} \alpha_{m, i, j },\\
    \gamma_j &= \frac{\exp(\beta_j)}{\sum_{k=0}^{i-1} \exp(\beta_k)}
\end{aligned}    
\end{equation}

\subsection{Feed-Forward Network}
We utilize Feed-Forward Network (FNN) module to enhance the learning ability of Mamba Block, $\mathrm{H}$ denotes the output of Mamba Block:
\begin{equation}
\label{eq:ffn}
\begin{aligned}
    \mathrm{FFN(H)}  &= \mathrm{GELU}(\mathrm{H}W^{(1)}+b^{(1)})W^{(2)} + b^{(2)}
\end{aligned}    
\end{equation}
where $W^{(1)}\in\mathbb{R}^{D \times 4D}, W^{(2)}\in\mathbb{R}^{4D \times D}, b^{(1)}\in\mathbb{R}^{4D}$ and $b^{(2)}\in\mathbb{R}^D$.

\subsection{Prediction Network}
In the prediction network, the output information f from Eq.~\ref{eq:ffn} is used to generate the final prediction probability:
\begin{equation}
\label{eq:predict}
\begin{aligned}
    p_t  &= Sigmoid(f*W+b) 
\end{aligned}    
\end{equation}
where $p_t$ represents the probability of correctly answering question $x_t$.

\subsection{Objective Function}
We select the cross-entropy log loss as our objective function, which compares the prediction $p_t$ with the actual answer $r_t$.

\begin{equation}
\label{eq:predict}
\begin{aligned}
    L = -\sum_t \left(r_t \log(p_t) + (1 - r_t) \log(1 - p_t)\right) + \lambda||\theta||^2
\end{aligned}    
\end{equation}
where $\lambda$ is the regularization hyperparameter and $\theta$ is the parameter of $\mu_{q_t}$ in Eq.~\ref{eq:QR}.
\section{Experiments}

\subsection{Experimental Setup}
\textbf{Datasets.}\quad 
We evaluated the effectiveness of Mamba4KT on three datasets: the ASSISTments 2012\footnote{https://sites.google.com/site/assistmentsdata/home/2012-13-school-data-with-affect.}, ASSISTments 2017\footnote{https://sites.google.com/view/assistmentsdatamining/dataset?authuser=0.} and Eedi\footnote{https://eedi.com/projects/neurips-education-challenge.}. We processed the data using the open source tool pykt. All three datasets include question ids. Detailed data in Table~\ref{table_dataset}.
\begin{itemize}
\item ASSISTments: Several datasets collected in the ASSISTments platform, an online tutoring system that provides math instruction and access services for students.
\item Eedi: Student responses to math questions collected over two school years (2018-2020) in the Eedi platform, a free homework and teaching platform for primary and secondary schools in the UK.
\end{itemize}

\begin{table}[H]
\caption{Statistics of the Three Datasets.}
\begin{tabular}{llll}
\hline
               & ASSISTments2012 & ASSISTments2017 & Eedi      \\ \hline
\#questions    & 53,070          & 3,162           & 948       \\
\#concepts     & 265             & 102             & 57        \\
\#Interactions & 2,709,647       & 942,814         & 1,399,520 \\
\#Avg.length   & 99              & 552             & 284       \\ \hline
\end{tabular}
\label{table_dataset}
\end{table}

\begin{table*}[h!]
\centering
\caption{Overall Performance Comparison on The Three Datasets.}
\resizebox{\textwidth}{!}{
\begin{tabular}{l|l|ccccccccc|cc}
\hline
Dataset                                               & Metric & \multicolumn{1}{l}{DKT} & \multicolumn{1}{l}{DKT-forget} & \multicolumn{1}{l}{HawkesKT} & \multicolumn{1}{l}{Deep-IRT} & \multicolumn{1}{l}{LPKT} & \multicolumn{1}{l}{SAKT} & \multicolumn{1}{l}{AKT} & \multicolumn{1}{l}{SAINT} & \multicolumn{1}{l|}{Folibikt} & \multicolumn{1}{l}{MambaKT} & \multicolumn{1}{l}{Improv.} \\ \hline
\multicolumn{1}{c|}{\multirow{5}{*}{ASSISTments2012}} & AUC    & 0.7303                  & 0.7333                         & 0.7483                       & 0.7276                       & 0.7720                   & 0.7073                   & \textbf{0.7795}         & 0.7082                    & {\ul 0.7786}                  & 0.7764                      & -0.40\%                     \\
\multicolumn{1}{c|}{}                                 & ACC    & 0.7335                  & 0.7355                         & 0.7403                       & 0.7327                       & 0.7527                   & 0.7242                   & \textbf{0.7574}         & 0.724                     & {\ul 0.756}                   & 0.7543                      & -0.41\%                     \\
\multicolumn{1}{c|}{}                                 & TT     & 279                     & 433                            & 296                          & 1618                         & 6596                     & 418                      & 4002                    & 2378                      & 4132                          & 612                         & 553.92\%                    \\
\multicolumn{1}{c|}{}                                 & IT     & 2.2                     & 2.3                            & 2.2                          & 4.1                          & 43.4                     & 1.8                      & 13.4                    & 7.0                       & 12.3                          & 4.2                         & 220.18\%                    \\
\multicolumn{1}{c|}{}                                 & GC     & 1.0                     & 0.9                            & 1.1                          & 4.9                          & 4.4                      & 0.9                      & 9.3                     & 5.1                       & 8.5                           & 2.0                         & 358.92\%                    \\ \hline
\multirow{5}{*}{ASSISTments2017}                      & AUC    & 0.7171                  & 0.7258                         & 0.7049                       & 0.7114                       & \textbf{0.7839}          & 0.6534                   & 0.7651                  & 0.6796                    & 0.7741                        & {\ul 0.7818}                & -0.27\%                     \\
                                                      & ACC    & 0.6862                  & 0.6914                         & 0.6866                       & 0.685                        & \textbf{0.7271}          & 0.6649                   & 0.7143                  & 0.6700                    & 0.7202                        & {\ul 0.7251}                & -0.28\%                     \\
                                                      & TT     & 68                      & 88                             & 68                           & 252                          & 2074                     & 94                       & 656                     & 386                       & 629                           & 167                         & 1141.92\%                   \\
                                                      & IT     & 0.4                     & 0.4                            & 0.4                          & 0.7                          & 7.0                      & 0.3                      & 2.2                     & 1.2                       & 2.1                           & 0.8                         & 786.38\%                    \\
                                                      & GC     & 0.7                     & 0.8                            & 0.7                          & 4.6                          & 1.5                      & 0.6                      & 9.0                     & 4.2                       & 8.2                           & 1.7                         & -13.13\%                    \\ \hline
\multirow{5}{*}{Eedi}                                 & AUC    & 0.7712                  & 0.7763                         & 0.6861                       & 0.7698                       & 0.7946                   & 0.7436                   & 0.7958                  & 0.7821                    & {\ul 0.8026}                  & \textbf{0.8033}             & 0.09\%                      \\
                                                      & ACC    & 0.7043                  & 0.7099                         & 0.6604                       & 0.7029                       & 0.7240                   & 0.6807                   & 0.7261                  & 0.7127                    & {\ul 0.7317}                  & \textbf{0.7328}             & 0.15\%                      \\
                                                      & TT     & 75                      & 97                             & 115                          & 397                          & 3739                     & 144                      & 1049                    & 662                       & 951                           & 194                         & 390.21\%                    \\
                                                      & IT     & 0.5                     & 0.7                            & 0.5                          & 1.0                          & 11.0                     & 0.5                      & 3.5                     & 1.9                       & 3.3                           & 1.1                         & 201.43\%                    \\
                                                      & GC     & 0.8                     & 0.8                            & 0.8                          & 4.6                          & 0.8                      & 0.6                      & 9.0                     & 4.1                       & 8.2                           & 1.7                         & 384.53\%                    \\ \hline
\end{tabular}
}
\label{all_experiment}
\end{table*}

\textbf{Baselines.}\quad 
We chose the models DKT, DKT-forget, and LPKT based on RNN structure, and the models SAKT, AKT, SAINT, Folibikt based on attention structure, as well as other representative models Deep-IRT, HawkesKT.

\textbf{Metircs.}\quad 
We use AUC and ACC, which are common metrics in knowledge tracing scenarios, as metrics for evaluating the predictive effectiveness of the model. We count three values as a measure of training and inference efficiency and resource utilization: total training time TT (in seconds), total testing time CT (in seconds), and the size of resources requested GC (in GB) during training. We provide sequence-level interpretable and exercise-level interpretable samples through visualization.

\textbf{Reproducibility Settings.}\quad 
We performed the training and testing based on the open source framework pyKT~\cite{pykt}. The batch size is fixed to 64, the size of hidden layer is chosen from $\{64, 128, 256\}$, and the setting of learning rate is chosen from $\{0.003, 0.002, 0.001, 0.0001\}$. The number of layers N of Mamba Blocks is set to 5. For all the methods, we use Adam optimizer. During the training process we use an early stopping strategy, if the values of the optimal AUC and ACC do not change within 10 epochs, we stop the training.

\subsection{Results and Analysis}
\noindent\textbf{Overall Results.} 
Table~\ref{all_experiment} shows the comparison results of Mamba4KT with the baseline models on the three datasets. From the experimental results, we can see that the prediction effect of Mamba4KT is comparable to the results of the optimal model on each dataset, but in terms of time and space consumption, it has a very significant improvement over the optimal model or the model based on the attention mechanism. Compared to current knowledge tracing models, we designed Mamba4KT with a simpler structure, which may be the reason why Mamba4KT did not achieve optimal results on the three datasets. On Assistments2017, the model of Mamba4KT outperforms the model based on the attention mechanism, which further suggests that Mamba4KT outperforms the model based on the attention mechanism on the dataset with longer average length. From the overall experimental results, the model based on RNN structure has a relatively large time consumption. And the model based on the attention mechanism has a relatively large space consumption. And Mamba4KT, which is based on mamba structure, has much lower time and space consumption than the model based on the attention mechanism, and also has much lower time consumption than the LPKT model based on RNN structure, and also has lower space consumption than the LPKT model on the relatively large scale dataset, Assistments2012. This further illustrates the feasibility of using Mamba for knowledge tracing tasks.

\begin{table}[h]
\centering
\caption{Results of The Ablation Study}
\resizebox{8cm}{!}{
\begin{tabular}{l|cc|cc|cc}
\hline
Dataset         & \multicolumn{2}{c|}{ASSISTments2012} & \multicolumn{2}{c|}{ASSISTments2017} & \multicolumn{2}{c}{Eedi}          \\ \hline
Metric          & AUC               & ACC              & AUC               & ACC              & AUC             & ACC             \\
SAKT            & 0.7073            & 0.7242           & 0.6534            & 0.6649           & 0.7436          & 0.6807          \\
Mamba4KT        & \textbf{0.7756}   & \textbf{0.7540}  & \textbf{0.7818}   & \textbf{0.7251}  & \textbf{0.8024} & \textbf{0.7308} \\ \hline
w/o FFN         & {\ul 0.7737}      & {\ul 0.7528}     & {\ul 0.7677}      & {\ul 0.7148}     & {\ul 0.8000}    & {\ul 0.7295}    \\
w/o Rasch       & 0.7546            & 0.7422           & 0.7297            & 0.6935           & 0.7928          & 0.7230          \\
w/o FFN + Rasch & 0.7224            & 0.7308           & 0.7224            & 0.6906           & 0.7682          & 0.7016          \\ \hline
\end{tabular}
}
\label{ablation}
\end{table}

\graphicspath{{pdfs/}}
\begin{figure}[h]
    \includegraphics[width=8cm]{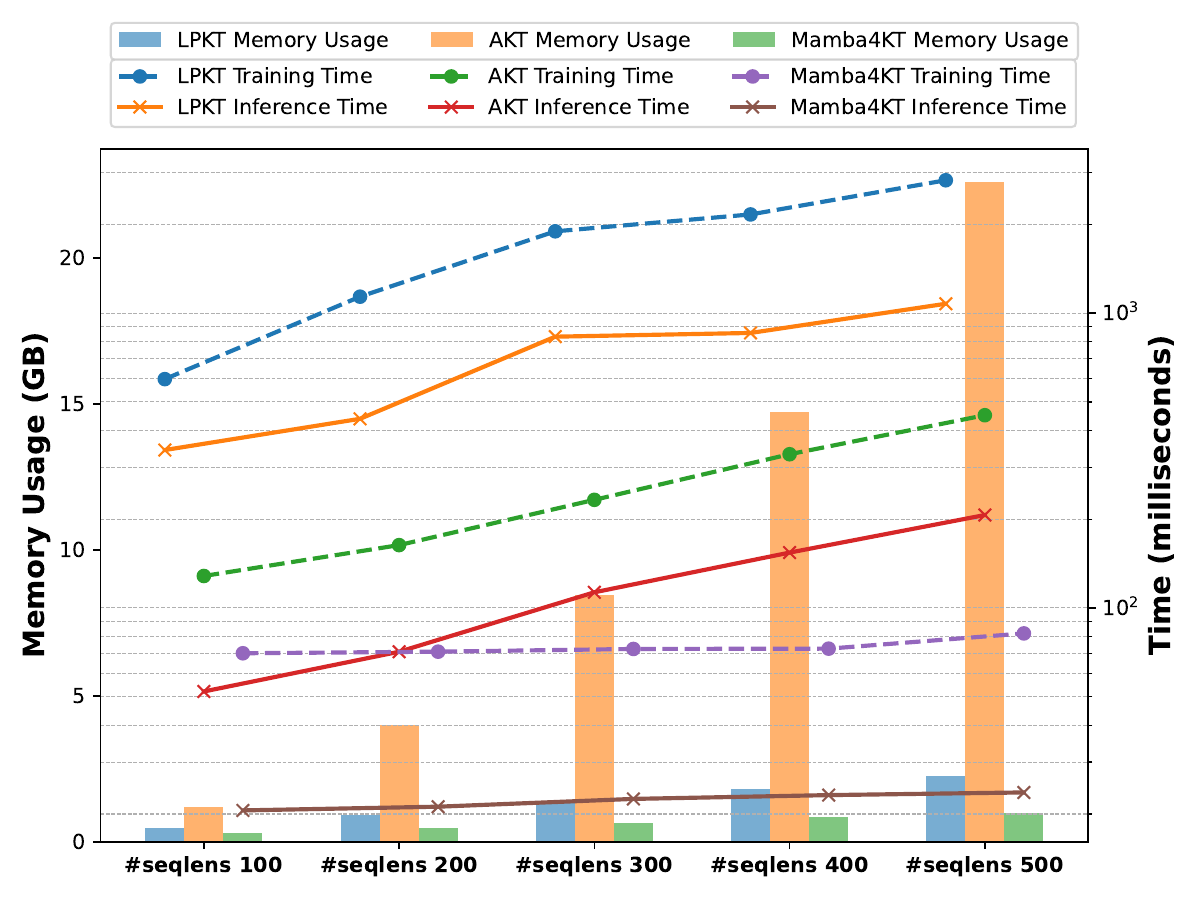}
    \caption{Comparison of model time and space consumption when dealing with sequences of different lengths.}
    \label{efficiency}
\end{figure}

\noindent\textbf{Ablation Study.}
Table~\ref{ablation} presents the results of the ablation experiments, which delineates the impact of various model components on the overall performance. The notation “w/o FFN” signifies the omission of the fully connected feed-forward network layer, while “w/o Rasch” indicates the exclusion of the Rasch component from the loss function ($\lambda$ in Eq.~\ref{eq:predict} is set to 0.). This comparison reveals that the Rasch component exerts a more pronounced influence on model prediction accuracy compared to FFN, echoing the findings from the AKT experiment, thus underscoring the versatility of attention-based strategies within the Mamba architecture.

Furthermore, the designation “w/o FFN+Rasch” indicates that only the Mamba Block is maintained, distinguishing it from the SAKT model. In the Mamba4KT, the Mamba structure is employed to encapsulate the inter-exercise information, whereas SAKT relies on a multi-head attention mechanism. The experimental data conclusively demonstrates that the Mamba module is comparable or even superior to models based on attention mechanisms in knowledge tracing tasks. Combined with the excellent performance of Mamba in terms of time and space consumption in Table~\ref{all_experiment}, it further illustrates its effectiveness and future development potential in the field of knowledge tracing.

\graphicspath{{pdfs/}}
\begin{figure*}[t!]
    \centering
    \includegraphics[width=18cm]{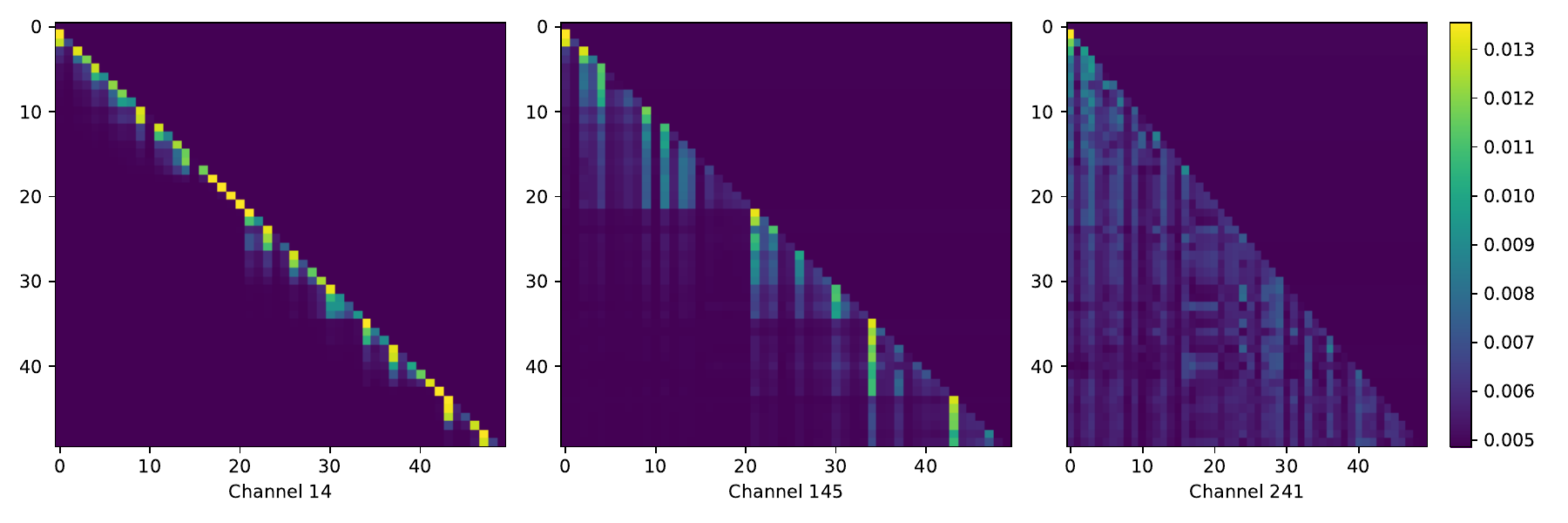}
    \caption{Sequence-level model interpretability.}
    \label{vis_channel}
\end{figure*}


\noindent\textbf{Time and Space Utilization.} 
Fig~\ref{efficiency} shows the training time, inference time, and GPU utilization of AKT, LPKT, and Mamba4KT on the Assistments2017 dataset with set sequence lengths from 100 to 500. As can be seen from the model efficiency and resource utilization, the training and inference time of LPKT increases rapidly as the sequence length increases, and LPKT is more than 10 times slower than mamba when seqlens is set to 500, while the memory usage of AKT is more than 20 times that of Mamba4KT. This comparison illustrates that a knowledge tracing model based on the mamba structure can be optimized in terms of the combined effect of time and space while ensuring prediction accuracy.

\subsection{Visualizing Attention Weights in Mamba Structures}

\noindent\textbf{Sequence-level interpretability.}
Figure 4 demonstrates the sequence-level interpretability of Mamba4KT based on the Assistments2017 dataset. We selected Channels 14, 145, and 241 from the 256 channels, each representing different types of information. Channels like Channel 14 capture short-range information between exercises approximately 4 steps apart, while Channels like Channel 145 capture information between exercises around 10 steps apart, and Channels like Channel 241 capture global relationships between exercises. This is similar to the multi-head attention mechanism in the AKT model, where different attention heads capture information at various levels. 
\graphicspath{{pdfs/}}
\begin{figure}[h]
    \includegraphics[width=9cm]{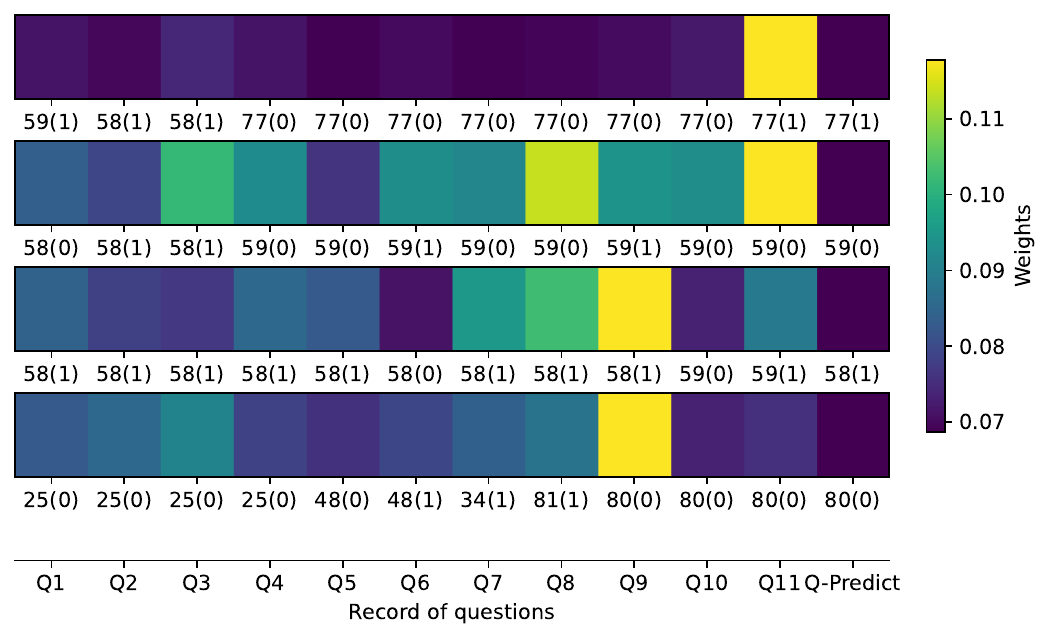}
    \caption{Exercise-level model interpretability. 59(1) indicates a correct answer to a question with a knowledge category of 59, and conversely 59(0) indicates an incorrect answer.}
    \label{vis_que}
\end{figure}

\noindent\textbf{Exercise-level interpretability.}
According to the Eq.~\ref{eq:exp_exe}, we aggregate the information from different channels to calculate the final attention weights. We present each subgraph using a top-down order.

The first two sub-figures reflect that Mamba4KT captures the information that learners' mastery of a particular type of knowledge increases or decreases.
The first subfigure reveals that students answered questions 4 to 10 incorrectly for knowledge point category 77, but correctly for Q11, suggesting a surge in their mastery of knowledge point 77 upon correctly answering Q11. The model successfully captured this trend and forecasted that Q11 exerted the most significant influence on the Q-predict.
The second subfigure shows that despite Q9 answering correctly, learners' knowledge level decreased after answering incorrectly to two consecutive questions with a knowledge point category of 59, with Q8 and Q11 playing a relatively large role in the predicted Q-Pridict.

The last two subfigures illustrate that Mamba4KT can differentiate learners' mastery on different categories of knowledge points.
The third subfigure shows that Q7, Q8, and Q9 with the same knowledge points have the greatest impact on the prediction results, while Q10 and Q11, which are closer but have different types of knowledge points, have less impact.
The fourth subfigure shows that even though Q6 through Q8 have correct answers, the impact weights are less than Q9 due to the different type of Q-Predict.

\section{Conclusion}
In this paper, we provide the first insight into the critical roles of time and space efficiency in the knowledge tracing task, and accordingly propose a novel knowledge tracing model, Mamba4KT, which not only takes into account the model's prediction effectiveness, time and space efficiency, but also integrates interpretability considerations, making it the first knowledge tracing model to combine these factors. Mamba4KT surpasses traditional RNN-based and attention-based models in terms of time and space efficiency through a unique mamba block structure that captures the intrinsic associations of learners' exercise sequences. In addition, the model incorporates advanced strategies such as rasch model embed-based embedding and FFN in current knowledge tracing models to significantly improve prediction accuracy.
We also apply the interpretability theory of mamba block in the visual domain to the knowledge tracing domain, providing a new interpretable approach to this task. Due to the relative simplicity of our model design, there is still room for improvement in prediction accuracy relative to the previously optimal model. Therefore, we plan to further investigate efficient and lightweight knowledge tracing models in order to achieve further performance improvements.


 






\end{document}